\documentclass[11pt]{article}
\usepackage{coling2018}
\usepackage{times}
\usepackage{url}
\usepackage{latexsym}
\usepackage{pdfpages}
\usepackage{amsmath}
\usepackage{tikz}
\usepackage{graphicx}
\usepackage{multirow}
\usepackage{setspace}
\usepackage{cases}
\usepackage{arydshln}
\usetikzlibrary{positioning,shapes,shadows,arrows,decorations.text}

\makeatletter

\newcommand{\Rmnum}[1]{\expandafter\@slowromancap\romannumeral #1@}
\makeatother

\usepackage{CJKutf8}

\slname{Chinese}

\title{Sequence-to-Sequence Data Augmentation for Dialogue Language Understanding}

\sltitle{\begin{CJK}{UTF8}{gbsn}对话语义理解的序列到序列数据增强\end{CJK}}

\author{Yutai Hou, Yijia Liu, Wanxiang Che*, Ting Liu\\
	Research Center for Social Computing and Information Retrieval \\
	Harbin Institute of Technology, China \\
	{\tt \{ythou, yjliu, car, tliu\}@ir.hit.edu.cn} \\\
}

\date{}

\begin{document}


\maketitle
\begin{abstract}
In this paper, we study the problem of data augmentation
for language understanding in task-oriented dialogue system.
In contrast to previous work which augments an utterance
without considering its relation with other utterances,
we propose a sequence-to-sequence generation
based data augmentation framework that leverages one utterance's 
same semantic alternatives in the training data.
A novel diversity rank is incorporated into the utterance representation
to make the model produce diverse utterances
and these diversely augmented utterances help to improve the language understanding module.
Experimental results on the Airline Travel Information System dataset
and a newly created semantic frame annotation on Stanford Multi-turn, Multi-domain
Dialogue Dataset show that our framework achieves significant improvements
of 6.38 and 10.04 F-scores respectively when only a training set of hundreds utterances is represented.
Case studies also confirm that our method generates diverse utterances.
\end{abstract}

\makesltitle
\begin{slabstract}
	\begin{CJK}{UTF8}{gbsn}
		在本文中，我们研究了面向任务的对话系统中语言理解模块的数据增强问题。相比之前的工作在生成新语句时不考虑语句间关系，我们利用训练数据中与一个语句具有相同语义的其他句子，提出了基于序列到序列生成的数据增强框架。我们创新地将多样性等级结合到话语表示中以使模型产生多样化的语句数据，而这些多样化的新语句有助于改善语言理解模块。在航空旅行信息系统数据集以及一个新标注的斯坦福多轮多域对话数据集上的实验结果表明，当训练集仅包含数百句语料时，我们的框架在F值上分别实现了6.38和10.04的显着提升。案例研究也证实我们的方法能够产生多样化的话语。
	\end{CJK}
\end{slabstract}

\section{Introduction}\label{sec:intro}

\blfootnote{
	%
	%
	%
	%
	 \hspace{-0.65cm}  
	 This work is licenced under a Creative Commons 
	 Attribution 4.0 International Licence.
	 Licence details:
	 \url{http://creativecommons.org/licenses/by/4.0/}
	%
	%
}
\blfootnote{
	\hspace{-0.65cm}
	* Email corresponding.
}

Language understanding (LU) is the initial and essential
component in the task-oriented dialogue system pipeline \cite{young2013pomdp}.
One challenge in building robust LU is 
to handle myriad ways in which users express demands.
This challenge becomes more serious
when switching to a new domain whose large-scale labeled data is
usually unreachable.
Insufficiency in training data makes LU 
vulnerable to unseen utterances which are syntactically different but semantically
related to the existing training data, and further harms the whole task-oriented dialogue system pipeline.

\textit{Data augmentation}, which enlarges the size of training data in machine learning systems,
is an effective solution to the data insufficiency problem.
Success has been achieved with data augmentation on a wide range of problems 
including computer vision \cite{NIPS2012_4824}, speech recognition \cite{DBLP:journals/corr/HannunCCCDEPSSCN14}, text classification \cite{zhang2015character}, and question answering \cite{fader-zettlemoyer-etzioni:2013:ACL2013}.
However, its application in the task-oriented dialogue system is
less studied.
\newcite{DBLP:conf/interspeech/KurataXZ16} presented the only work we know
that tried to augment data for LU.
In their paper, an encoder-decoder is learned to reconstruct
the utterances in the training data.
During the augmenting process, 
the encoder's output hidden states are randomly perturbed
to yield different utterances.

The work of \newcite{DBLP:conf/interspeech/KurataXZ16}
augments one single utterance by adding noise without considering its relation
with other utterances.
Besides theirs, there are also works which explicitly consider
the paraphrasing relations between instances that share the same output.
These works achieve improvements on tasks like text classification and question answering.
Paraphrasing techniques including word-level substitution \cite{zhang2015character,wang2015s},
hand-crafted rules generation \cite{fader-zettlemoyer-etzioni:2013:ACL2013,jia-liang:2016:P16-1},
and grammar-tree generation \cite{narayan-reddy-cohen:2016:INLG} have been explored.
Compared with these work, \newcite{DBLP:conf/interspeech/KurataXZ16} has 
the advantage of fully data-driven method and can easily switch to new domain without too much domain-specific knowledge,
but doesn't make use of the relations  between instances within the training data.

In this paper, we study the problem of data augmentation for LU 
and propose a novel data-driven framework that models relations between utterances
of the same semantic frame in the training data.
A sequence-to-sequence (seq2seq, Sutskever et al. 2014)
model lies in the core of our framework which takes
a delexicalised utterance and generates its lexical and syntactical alternatives.
To further encourage diverse generation, 
we incorporate a novel \textit{diversity rank} into the utterance representation.
When training the seq2seq model, 
the diversity rank is also used to filter the over-alike pairs of alternatives.
These approaches lead to diversely augmented data that significantly
improves the LU performance in the domains that labeled data is scarce.

We conduct experiments on the
Airline Travel Information System dataset (ATIS, Price 1990\nocite{Price:1990:ESL:116580.116612})
along with a newly annotated layer of slot filling over the
Stanford Multi-turn, Multi-domain Dialogue Dataset \cite{eric2017key}.\footnote{abbreviated as \textit{Stanford dialogue dataset} henceforth.}
On the small proportion of ATIS which contains 129 utterances,
our method outperforms the baseline
by a 6.38 F-score on slot filling.
On the medium proportion, this improvement is 2.87.
Similar trends are witnessed on our LU annotation over Stanford dialogue dataset
which the average improvement on three new domains is 10.04 on 100 utterances and 0.47 on 500 utterances.
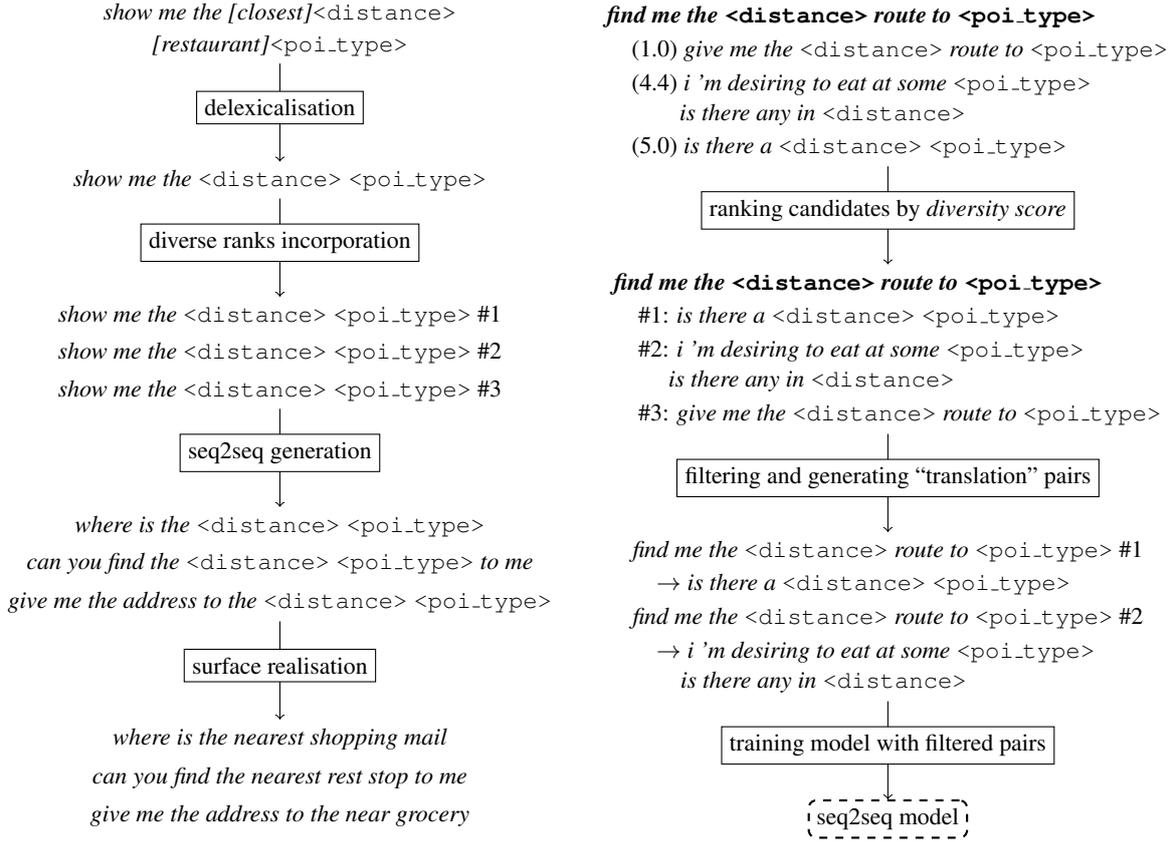
\begin{figure}[t]
	\small
	\centering
	\begin{tikzpicture}[node distance=1.3cm]
	\node (p1) [align=center] {\textit{show me the [closest]}\texttt{<distance>} \\ [0.5ex] \textit{[restaurant]}\texttt{<poi\_type>}};
	\node (p2) [below=of p1] {\textit{show me the} \texttt{<distance>}  \texttt{<poi\_type>}};
	\node (p3) [below=of p2, rectangle split, rectangle split parts=3] {
		\nodepart{one}
		\textit{show me the} \texttt{<distance>} \texttt{<poi\_type>} \#1
		\nodepart{two}
		\textit{show me the} \texttt{<distance>} \texttt{<poi\_type>} \#2
		\nodepart{three}
		\textit{show me the} \texttt{<distance>} \texttt{<poi\_type>} \#3
	};
	\node (p4) [below=of p3, rectangle split, rectangle split parts=3] {
		\nodepart{one}
		\textit{where is the} \texttt{<distance>} \texttt{<poi\_type>}
		\nodepart{two}
		\textit{can you find the} \texttt{<distance>} \texttt{<poi\_type>} \textit{to me}
		\nodepart{three}
		\textit{give me the address to the} \texttt{<distance>} \texttt{<poi\_type>}
	};
	\node (p5) [below=of p4, rectangle split, rectangle split parts=3] {
		\nodepart{one}
		\textit{where is the nearest shopping mail}
		\nodepart{two}
		\textit{can you find the nearest rest stop to me}
		\nodepart{three}
		\textit{give me the address to the near grocery}
	};
	
	\draw[->] (p1.south) -- (p2.north) node[pos=0.45, fill=white, draw=black] {delexicalisation};
	\draw[->] (p2.south) -- (p3.north) node[pos=0.45, fill=white, draw=black] {diverse ranks incorporation};
	\draw[->] (p3.south) -- (p4.north) node[pos=0.45, fill=white, draw=black] {seq2seq generation};
	\draw[->] (p4.south) -- (p5.north) node[pos=0.45, fill=white, draw=black] {surface realisation};
	
	\node (t1) [align=left] at (8cm, -0.7cm) {
		\textbf{\textit{find me the} \texttt{<distance>} \textit{route to} \texttt{<poi\_type>}} \\ [0.3em]
		\quad (1.0) \textit{give me the} \texttt{<distance>} \textit{route to} \texttt{<poi\_type>} \\ [0.3em]
		\quad (4.4) \textit{i 'm desiring to eat at some} \texttt{<poi\_type>} \\ [0.15em] \quad \qquad \textit{is there any in} \texttt{<distance>} \\ [0.3em]
		\quad (5.0) \textit{is there a} \texttt{<distance>} \texttt{<poi\_type>}
	};
	\node (t2) [below=of t1, align=left] {
		\textbf{\textit{find me the} \texttt{<distance>} \textit{route to} \texttt{<poi\_type>}} \\ [0.3em]
		\quad \#1: \textit{is there a} \texttt{<distance>} \texttt{<poi\_type>} \\ [0.3em]
		\quad \#2: \textit{i 'm desiring to eat at some} \texttt{<poi\_type>} \\ [0.15em] \ \qquad \textit{is there any in} \texttt{<distance>} \\ [0.3em]
		\quad \#3: \textit{give me the} \texttt{<distance>} \textit{route to} \texttt{<poi\_type>}
	};
	\node (t3) [below=of t2, align=left] {
		\textit{find me the} \texttt{<distance>} \textit{route to} \texttt{<poi\_type>} \#1 \\ [0.3em]
		\quad $\rightarrow$ \textit{is there a} \texttt{<distance>} \texttt{<poi\_type>} \\ [0.3em]
		\textit{find me the} \texttt{<distance>} \textit{route to} \texttt{<poi\_type>} \#2 \\ [0.3em]
		\quad $\rightarrow$ \textit{i 'm desiring to eat at some} \texttt{<poi\_type>} \\ [0.15em] \qquad \textit{is there any in} \texttt{<distance>}
	};
	\node (t4) [below=of t3, draw=black,dashed,thick,, rounded corners=0.1cm] {seq2seq model};
	\draw[->] (t1.south) -- (t2.north)  node[pos=0.45, fill=white, draw=black] {ranking candidates by \textit{diversity score}};
	\draw[->] (t2.south) -- (t3.north) node[pos=0.45, fill=white, draw=black] {filtering and generating ``translation'' pairs};
	\draw[->] (t3.south) -- (t4.north) node[pos=0.45, fill=white, draw=black] {training model with filtered pairs};
	
	\end{tikzpicture}
	\caption{The workflow of our framework. 
		The left part shows the augmenting process 
		and the right part shows the training instance generation process for our seq2seq model.
		$\mathbf{u} \rightarrow \mathbf{u}'$ marks that $\mathbf{u}$ can be augmented into $\mathbf{u}'$.}\label{fig:workflow}
\end{figure}

The major contributions of this paper include:
\begin{itemize}
	\item We propose a data augmentation framework for LU (\S\ref{sec:overview}) using the seq2seq model.
	 A novel diversity rank (\S\ref{sec:ranks}) is used to encourage our seq2seq model to generate diverse utterances
	 both in the augmentation and training (\S\ref{sec:learn}).
	\item We conduct experiments on the ATIS and Stanford dialogue dataset (\S\ref{sec:exp}).
	 Experimental results show our augmentation can effectively enlarge the training data and improve LU performance
	 by a large margin when only a small size of training data is presented.
	 Case studies also confirm that our method generates diverse utterances compared to the results from previous work.
\end{itemize}

We release our code at:

 \url{https://github.com/AtmaHou/Seq2SeqDataAugmentationForLU}.

\section{Overview of the Approach}\label{sec:overview}
\paragraph{Notion and Problem Description.}
In this paper, we study the data augmentation for language understanding (LU),
which maps a natural language utterance into its semantic frames.
We focus on \textit{slot filling} and follow previous works \cite{225939}
by treating it as a sequence
classification in which semantic class labels (\textit{slot types}) are assigned to
contiguous sequences of words indicating these sequences are corresponding \textit{slot values}.
In this paper, we use the bidirectional long short term memory (BiLSTM) 
for slot labeling (tagging)
as previous works did \cite{DBLP:conf/interspeech/MesnilHDB13,7078572,kurata-EtAl:2016:EMNLP2016}.

We formalize the data augmentation for LU as given a 
natural language utterance $\mathbf{u}$ and its semantic frame $\mathbf{s}$,
we generate a set of new utterances with corresponding semantic frames.
During the augmenting process, we go through the whole training data
$D=\{(\mathbf{u}_{i}, \mathbf{s}_{i})\}_{i=1}^N$.
For each training instance $(\mathbf{u}_i, \mathbf{s}_i)$, we expand it to
a set of instances $\{(\mathbf{u}_i^k, \mathbf{s}_i^{k})\}_k$
and use the union of the expanded instances as new data to train the LU module.

In the training phase, we define the cluster of semantic frame $\mathbf{s}$
as $C_{\mathbf{s}}=
\{(\mathbf{u'}, \mathbf{s'}) \mid (\mathbf{u'}, \mathbf{s'}) \in D \land \mathbf{s'} = \mathbf{s}\}$.
For one utterance $\mathbf{u}$ and its semantic frame $\mathbf{s}$, 
each utterance $\mathbf{u}' \in C_{\mathbf{s}} / \{(\mathbf{u}, \mathbf{s})\}$ is considered as the alternative expression and augmentation of $\mathbf{u}$.
We use $\mathbf{u} \rightarrow \mathbf{u}'$ to mark this relation.

To achieve the goal of generating variant utterances under the same semantic frames,
We break down the problem into 
first converting the input utterance $\mathbf{u}$ into its delexicalised form $\mathbf{d}$,
and then generating the delexicalised variances of $\mathbf{d}$ with 
a seq2seq model.
Finally, surface realization is carried out to convert the delexicalised form into the raw utterance.
The left part of Figure \ref{fig:workflow} shows the workflow of our augmenting process.

\paragraph{Delexicalisation.} 

When given the raw utterance and its semantic frames
associated with certain segments of the utterance,
we can easily delexicalise the utterance by replacing the corresponding segments
with the semantic frame label.
For example, when given the 4th word in ``\textit{show me the closest restaurant}''
as a \texttt{<distance>} slot type and 5th word as \texttt{<poi\_type>} slot type,
its delexicalized form ``\textit{show me the} \texttt{<distance>} \texttt{<poi\_type>}''
is straight-forward to achieve.

In the task-oriented dialogue system, slot values usually consist of various entity names and are very sparse.
Delexicalisation reduces the size of vocabulary and makes the model
focus more on generating variant ways of expressing demands.
What's more, the semantic frames can be directly derived from
the delexicalised generation and used for training the LU module.

\paragraph{Incorporating Diversity Ranks into Utterance Representations.}
Considering the example in the
right part of Figure \ref{fig:workflow},
``\textit{is there  a} \texttt{<distance>} \texttt{<poi\_type>}'' is more diverse than
``\textit{give me the} \texttt{<distance>} \textit{route to} \texttt{<poi\_type>}''
when compared with ``\textit{find me the} \texttt{<distance>} \textit{route to} \texttt{<poi\_type>}''.
This example shows that for utterance $\mathbf{u}$ with semantic frame $\mathbf{s}$,
its alternatives expressions can have different ranks in diversity.
To consider the ranking information, we
compile the \textit{diversity rank} as an additional information
into the utterance representation.
By setting it to a higher rank, we aim to generate input utterance's diverse augmentation,
and by setting it to lower, a similar utterance should be generated.
We will discuss the details of how to compute the ranks during training 
and how to decide the effective numbers of ranks during testing in Section \ref{sec:ranks}. 

\paragraph{Data Augmentation as Seq2Seq Generation.}
When given the delexicalised input utterance $\mathbf{d}$ and the specified diverse rank $k$,
we use the standard seq2seq model to generate the alternative delexicalised
utterance $\mathbf{d'}$.
In our seq2seq model, we append \#$k$ to the end of the input utterances and
the model is formalized as
\[
p(\mathbf{d'} \mid \mathbf{d}, k) = \prod_{t}{p(d'_t \mid d_1, ..., d_n, \text{\#}k, d'_1, ...,d'_{t-1})}
\]
where \(n\) is the number of words for the input utterance \(\mathbf{d}\). 

In this paper, we follow the seq2seq model for neural machine translation and
use the \textit{input-feeding} network in \cite{luong-pham-manning:2015:EMNLP}
with attention as our seq2seq model.
During testing, we use beam search with beam size of 10
to yield more than one translation following \newcite{gimpel-EtAl:2013:EMNLP} and \newcite{DBLP:journals/corr/VijayakumarCSSL16}.

To train the seq2seq model, our basic assumption is that if $\mathbf{d}$ and $\mathbf{d'}$
contain the same semantic frames, they can be generated from each other.
Generally, we assume each pair of delexicalised utterances
in the cluster $C_\mathbf{s}$ makes a pair of generation.
However, it's nontrivial to assign \textit{diverse ranks} to training data.
What's more, to prevent the model from just producing produce lexical paraphrases
(like ``\textit{show me}'' to ``\textit{give me}''),
we propose to also consider the diversities when generating training
translations for the  seq2seq model.
We will talk about the details in Section \ref{sec:learn}.

%
%
%

\paragraph{Surface Realisation.}

Till now, we have achieved the lexically and syntactically different utterances in their delexicalized forms.
We would like to bridge these utterances to their lexicalized forms and
surface realisation is employed as the final step of our approach.


In this paper, the surface realisation is performed by
replacing the slot type in the delexicalised form
with its slot value.
The mapping from slot type to its set of slot values 
(e.g. from \texttt{<poi\_type>} to \{\textit{hospital}, \textit{restaurant}\})
is collected on the training data.
Somehow, it's nontrivial to just do the replacement
because one slot value doesn't fit its slot type in any context.
Taking the utterance in Figure \ref{fig:workflow} for example,
in the delexicalised utterance ``\textit{i 'm desiring to eat at some} \texttt{<poi\_type>} \textit{is there any in} \texttt{<distance>}'', `hospital' doesn't fit in the \texttt{<poi\_type>}
because `hospital' isn't the place intended for a meal.
To make the surface realisation more reasonable, we
build the mapping with consideration of the context
and use slot type along with its surrounding 5 words
as the key in the mapping.

During surface realisation for an utterance,
we first extract the slot type and its context.
Then we use this to get all its slot values.
If the slot type under certain context is not presented in
the mapping, we use the one with the most similar context
in the sense of \textit{edit distance}.
If more than one slot values present, we randomly pick a slot value.

 


\section{\textit{Diversity Ranks} in Utterance Representations}\label{sec:ranks}

The major motivation of this paper is to encourage diverse generation.
To accomplish this motivation, we propose a criterion named \textit{diversity rank}
to model the diversities.
During augmenting the data, for an instance $(\mathbf{u}, \mathbf{s})$
we generate the 
delexicalised utterance at rank from 1 to $N_{\mathbf{s}}$, where $N_{\mathbf{s}}$
is a number governed by the semantic frame $\mathbf{s}$ and calculated as 
$|| C_{\mathbf{s}} || / 2$,
which is the half size of
the instances in $D$ that have the semantic frame $\mathbf{s}$.

During training the seq2seq model with diversity rank,
for one instance $(\mathbf{u}, \mathbf{s})$, we first collect $C_\mathbf{s}$,
then rank each instance $(\mathbf{u'}, \mathbf{s}) \in C_\mathbf{s} / \{(\mathbf{u}, \mathbf{s})\}$
by its \textit{diversity score} against $\mathbf{u}$.
In this paper, 
the diversity score of an utterance pair $(\mathbf{u}, \mathbf{u'})$ is
calculated by both considering the \textit{edit distance}
and a \textit{length difference penalty} (LDP) as:
\begin{equation}\label{eq:score}
\textsc{score}(\mathbf{u},\mathbf{u'}) = \textsc{EditDistance}(\mathbf{u},\mathbf{u'}) \times \textsc{LDP}(\mathbf{u},\mathbf{u'})
\end{equation}
where LDP is defined as $\text{LDP}(\mathbf{u},\mathbf{u'}) = e^{-\frac{| ||\mathbf{u} || - || {\mathbf{u'}}|| |}{ || \mathbf{u} || }}$.
After obtaining the ranks over the utterances $\mathbf{u'}$,
we directly incorporate the rank value as an additional last token for 
the seq2seq model.

We note that using the LDP reduces the impact of differences in length
and makes the score paying more attention to the lexical and syntactical difference.
For example, the first block of right part of Figure \ref{fig:workflow} shows the diversity scores of three different
utterances.
Although the utterance ``\textit{i 'm desiring to eat at some} \texttt{<poi\_type>} 
\textit{is there any in} \texttt{<distance>}'' presents larger \textit{edit distance} (12 in this case)
than that of ``\textit{is there a} \texttt{<distance>} \texttt{<poi\_type>}'' (5 in this case),
the final score is penalized to 4.4 because the length difference.

In our method, the diversity rank can be treated as an utterance-independent controller
for the diversity of target generation.

\section{Filtering the Alike Instances}\label{sec:learn}

To learn the seq2seq model, it's straight-forward to use each pair of utterances
in $C_\mathbf{s}$ as training data for
the model. 
However, the goal of our paper is to generate diverse augmented data and
the usefulness of less diverse pair (like \textit{give me the} \texttt{<distance>} \textit{route to} \texttt{<poi\_type>}
and \textit{find me the} \texttt{<distance>} \textit{route to} \texttt{<poi\_type>} in Figure \ref{fig:workflow}) is arguable.

In this paper, we propose to filter the less diverse pairs when training the seq2seq model.
Again, we make use of the ranks derived by the diversity scores and for an utterance $\mathbf{u}$
only
the most diverse half of the translations $\mathbf{u} \rightarrow \mathbf{u'}$ are used
to train the seq2seq model
and the training data can be formalized as
\[
D_{\text{seq2seq}} = \bigcup_{(\mathbf{u}, \mathbf{s}) \in D} \{ \mathbf{u}, \textsc{Rank}(\mathbf{u}, \mathbf{u}') \rightarrow \mathbf{u}' \mid \mathbf{u}' \in C_\mathbf{s}, \textsc{Rank}(\mathbf{u}, \mathbf{u}') \ge ||C_{\mathbf{s}}|| / 2\}
\]
After filtering the less diverse pairs, we use $D_{\text{seq2seq}}$
to train the seq2seq model.

In this section, we revisit the role of our diversity ranks in the learning perspective.
Since we consider the utterance in cluster $C_\mathbf{s}$ as translation to
each other, without the \textsc{Rank} value, one utterance
can simultaneously translate to different utterances
in the training data.
It increases the ambiguities in learning the seq2seq model
and even makes it intractable.
With the \textsc{Rank} value, such ambiguities are resolved
because each pair of the training data is expanded with
a unique value.

\section{Experiments}\label{sec:exp}

\subsection{Settings}

\paragraph{Dataset.}

	
In this paper, we conduct our experiments on the ATIS dataset which is
extensively used for LU \cite{DBLP:conf/interspeech/MesnilHDB13,DBLP:journals/taslp/MesnilDYBDHHHTY15,chen2016syntax}.
The ATIS dataset contains 4978 training utterances from Class A
training data in the ATIS-2 and ATIS-3 corpus, while the test contains 893
utterances from the ATIS-3 Nov93 and Dec94 datasets.
The size of the training data is relatively large for LU in a single domain.
To simulate the data insufficient situations, we follow \newcite{chen2016syntax},
and also evaluate our model on two small proportions of the training data
which is \textit{small} (1/40 of the original training set with 129 instances) proportion and
\textit{medium} (1/10 of the original training set with 515 instances).
In all the experiments, a development set of 500 instances is used.

\begin{table}[t]
	\centering
	\begin{tabular}{lccc}
		\hline
		& Navigation & Scheduling & Weather \\ 
		\hline
		\# of training utterances & 500 & 500 & 500 \\
		\# of devel. utterances & 321 & 201& 262\\
		\# of test utterances & 337 & 212 & 271\\
		\hdashline
		Kappa & 0.68 & 0.92 & 0.90 \\
		Agreement & 85.05 & 90.75 & 95.99\\ 
		\hline
	\end{tabular}
	\caption{Statistics for our annotation.
	}
	\label{tbl:anno}
\end{table}

To test our model on new domains beyond ATIS,
we also create a new LU annotation over the Stanford dialogue dataset \cite{eric2017key}.
We use the same data split as \newcite{eric2017key}
and annotate the full test sets for the three domains (\textit{navigation},
\textit{scheduling}, and \textit{weather}) along with a small training set of
500 utterances.
The Stanford dialogue dataset provides semantic frames (\textit{slot})
for each utterance but doesn't associate the semantic class of the slot with corresponding
segment in the utterance.
Our annotation focus on assigning the slot to its corresponding segment.
During the annotation, each dialogue was processed by two annotators.
Data statistics, Kappa value \cite{snow2008cheap}, and inner annotator agreement
measured by F-score
on the three domains are shown in Table \ref{tbl:anno}.

\paragraph{Evaluation.} We evaluate our data augmentation's effect on LU
with F-score. \texttt{conlleval} is used in the same way with previous works \cite{DBLP:conf/interspeech/MesnilHDB13,DBLP:journals/taslp/MesnilDYBDHHHTY15,chen2016syntax}.

\paragraph{Implementation.}
We use OpenNMT \cite{klein-EtAl:2017:ACL-2017-System-Demonstrations}
as the implementation of our seq2seq model.
We set the number of layers in LSTM as 2 and the size of hidden states as 500.
Utterances that are longer than 50 are truncated.
We adopt the same training setting as \newcite{luong-pham-manning:2015:EMNLP}
and use Adam \cite{kingma2014adam} to train the seq2seq model.
Learning rate is halved when perplexity on the development set doesn't decrease.
During generation, we replace the model-yielded unknown token (\textit{unk}) with the source word
that has the highest attention score.

For the slot tagging model,  we set both the
dimension for word embedding and the size of hidden state to 100.
We also vary dropout rate in \{0, 0.1, 0.2\} considering
its regularization power on small size of data. 
The batch size is set to 16 in all the experiments.
Best hyperparameter settings are determined on the development set.
GloVe embedding \cite{pennington2014glove}
is used to initialize the word embedding in the model.
Adam with the suggested settings in \newcite{kingma2014adam}
is used to train the parameters.

\newcite{reimers-gurevych:2017:EMNLP2017} 
pointed out that neural network training is 
nondeterministic and depends on the seed for the random
number generator.
We witness dramatic changes of the slot tagging performance using
different random seeds.
To control for this effect, we take their suggestions and report the average of 5 differently-seeded runs.



\subsection{Results on ATIS}

\begin{table}[t]
	\centering
%
	\begin{tabular}{l|r@{.}l r@{.}l r@{.}l}
	\hline
	Model & \multicolumn{2}{c}{small} & \multicolumn{2}{c}{medium} & \multicolumn{2}{c}{full} \\ 
	& \multicolumn{2}{c}{129} & \multicolumn{2}{c}{515} & \multicolumn{2}{c}{4,478} \\
	\hline
	Baseline & 67 & 33** & 85 & 85** & 94 & 93* \\
	Ours  & 73 & 71 & 88 & 72 & 94 & 82\\
	
	Re-implementation of \newcite{DBLP:conf/interspeech/KurataXZ16} & 67 & 93** & 87 & 34** & 94 & 61** \\
	\hdashline
	Model-1 Additive \cite{DBLP:conf/interspeech/KurataXZ16} & \multicolumn{2}{c}{-} & \multicolumn{2}{c}{-} & 95 & 08 \\
	K-SAN syntax \cite{chen2016syntax} & 74 & 35 & 88 & 40 & 95 & 00 \\
	Model-\Rmnum{3}  \cite{zhai2017neural} & \multicolumn{2}{c}{-} & \multicolumn{2}{c}{-} & 95 & 86 \\
	\hline
\end{tabular}
	\caption{The results on the ATIS dataset.
		The first block shows the results from our implementation and
		the second block is drawn from the papers of previous works.
		Here we use * to indicate that the difference 
		between the model and Ours is statistically significant under t-test
		(** for p-value threshold as 0.05 and * for threshold as 0.1)
		. 
	}
	\label{tbl:atis-res}
\end{table}

Table \ref{tbl:atis-res} shows the slot tagging results on the ATIS dataset.
Our baseline model is the vanilla BiLSTM slot tagger
and our augmented slot tagger use the same architecture but is trained with the augmented
data generated by our method.
Compared with the vanilla tagger baseline, our augmentation method significantly improves
the LU performance by a 6.38 F-score  on the \textit{small}
proportion and a 2.02 F-score on the \textit{medium} proportion.
The improvements show the effectiveness of our augmentation method
in the data-insufficient scenario.
On the full data, our augmentation slightly lags the baseline.
We address this to the fact that full ATIS is large enough for LU on a single domain
and our augmentation introduce some noise.

To compare with the previous augmentation work from \newcite{DBLP:conf/interspeech/KurataXZ16},
we re-implemented their \textit{model-1 additive} model using the suggested settings in their paper.
The results on the \textit{small}, \textit{medium}, and \textit{full} proportions are shown
in the third row of Table \ref{tbl:atis-res}.
On all the proportions, our augmentation method outperforms theirs
and the differences are significant on \textit{small} and \textit{medium}.
Since their model relies on learning a seq2seq model to reconstruct the input utterances,
it's usually difficult to train a reasonable model on very small data due to sparsity.
Our method mitigates this by both generating on the delexicalised utterances
and learning the generation model from pairs of utterances that share same semantic frame
which enlarge the size of data for us to train the model.
\begin{table}[t]
	\centering
	\begin{tabular}{ll|ccc}
		\hline
		\# utterances & Model & Navigation & Scheduling & Weather \\
		\hline
		100 & Baseline  & 59.93 & 68.29 & 82.43 \\
		& Ours & 72.91 & 77.30 & 90.55 \\
		\hline
		\hline
		500 & Baseline & 78.99 & 86.05 & 93.68 \\
		& Ours & 78.46 & 87.67 & 94.01 \\
		\hline
	\end{tabular}
	\caption{The results on Stanford dialogue dataset.}\label{tbl:stanford-res}
\end{table}
We also compare our model with the syntax version of K-SAN \cite{chen2016syntax}
without joint training from intent annotation.
We see that our augmented tagger lags their 
syntax-parsing-enhanced model by a 0.64 F-score on small proportion and
outperforms theirs by a 0.32 F-score on medium proportion.
But considering the training data is sampled with
different random seeds between our work and theirs,
these results are not directly comparable.
At last, we show the \cite{zhai2017neural} as state-of-art results on ATIS dataset, 
which views slot filling task as sequence chunking problem.
As we focus data augmentation for sequence labeling task rather than chunking, 
this result is not directly comparable to ours.
Besides, K-SAN \cite{chen2016syntax} and \cite{zhai2017neural} are not 
data augmentation methods, 
we included their results to show that our augmentation method is reasonably good
The basic trend shows that our augmentation
can be used as an alternative to the LU model leveraging rich syntactic information.


\subsection{Results on Stanford Dialogue Dataset}

The results for Stanford dialogue dataset are shown in Table \ref{tbl:stanford-res}.
Similar trend as the ATIS experiments is witnessed in which the augmentation
improves the LU performance.
The average improvement on the training data with 100 utterances is 10.04,
and the number is 0.47 for that with 500 utterances.
Considering that only fewer than 350 utterances present in the test set
in all these domains, these improvements are reasonable.
Besides, similar to the ATIS results, the margin of improvements is larger for the smaller training set.

An advantage of our method is that it's purely data-driven. 
Only a mapping from slot type context to slot values is required and it can be
constructed from the training data.
It's easy for our method to switch to new domains and our results on the Stanford
dialogue dataset confirms this.
	
	
	
	
\subsection{Analysis}
%
%

\paragraph{Ablation.}
\begin{table}[t]
	\centering
	\begin{tabular}{l|r@{.}lcc}
		\hline
		Model & \multicolumn{2}{c}{F-score} & \# new & max. ED \\
		\hline
		Ours & 88 & 72 & 301 & 3.18 \\
		\quad - seq2seq generation & \ -0 & 84** & 0 & 0 \\
		\quad - diversity ranks & \ -0 & 40*  & 163 & 2.42 \\
		\quad - filtering &\ -0 & 38  & 870 & 2.86 \\
		\hline
	\end{tabular}
\caption{The result of the ablation test. 
	\textit{\# new} marks the number of newly generated delexicalised utterances. 
	\textit{max. ED} marks the averaged maximum edit distances.
	Here we use * to indicate that the result is statistically significant under t-test
	(** for p-value threshold as 0.05 and * for threshold as 0.1)
	By removing the seq2seq generation from our method, no delexicalised utterance will be generated
	so the \textit{max. ED} cell is 0. 
}\label{tbl:ablation}
\end{table}

\begin{figure}[t]
	\small
	\centering\includegraphics[width=0.73\columnwidth]{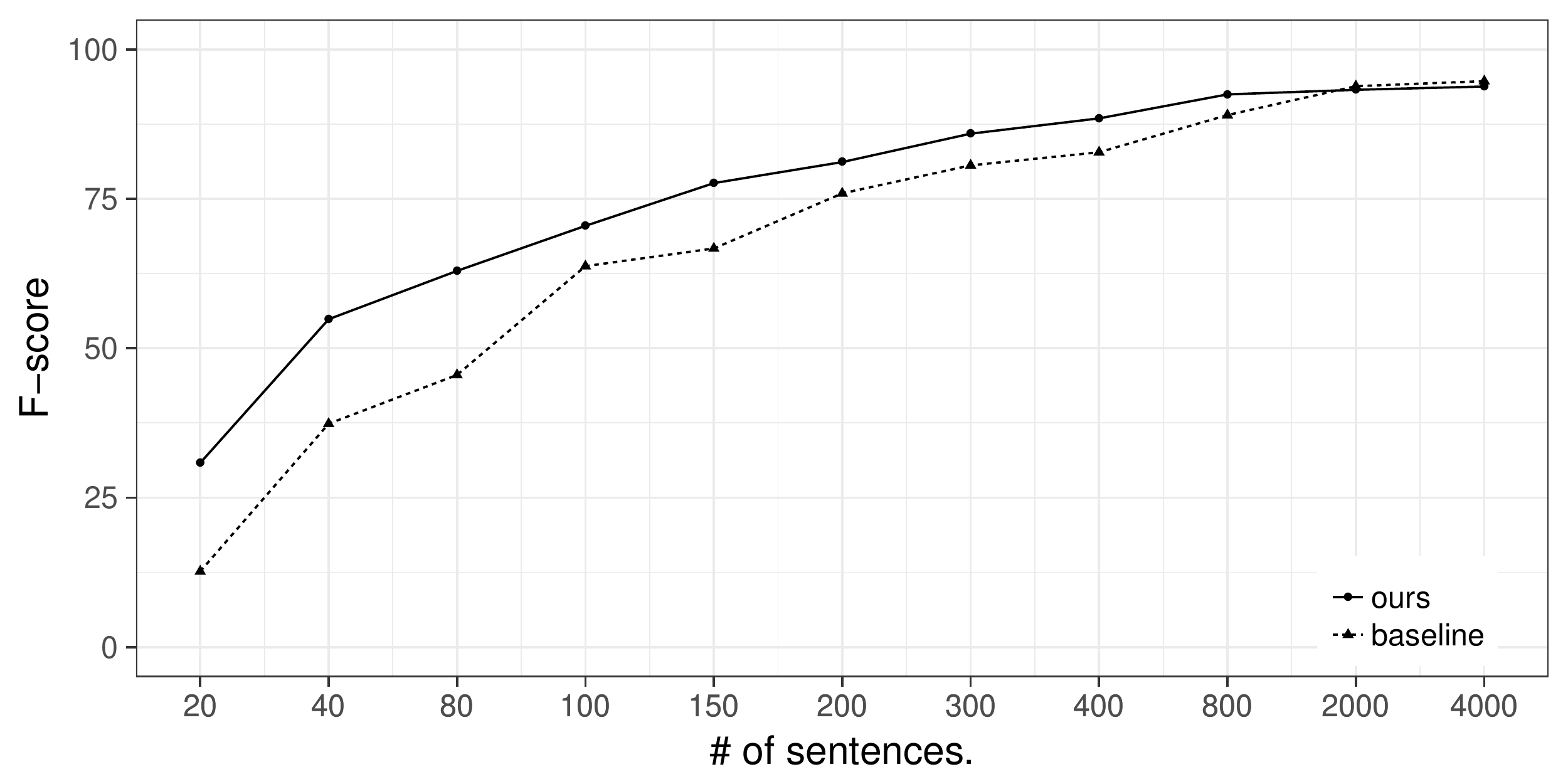}
	\caption{Our method's performances on the ATIS training data of different sizes.}\label{fig:data-vol}
\end{figure}

To get further understanding of each component in our method,
we conduct ablation on the \textit{medium} proportion,
Each of the three parts of our method is removed respectively, 
including the \textit{seq2seq generation}, \textit{diversity ranks}, and \textit{filtering}.
In addition to evaluate the model's performance with F-score,
we also examine the augmented data by the number of
newly generated delexicalised utterances and the maximum edit distances against the rest of instances.\footnote{This number is normalized by the total number of utterances.}
The results are shown in Table \ref{tbl:ablation}.

For our method without \textit{seq2seq generation}, we only conduct surface realisation on
the delexicalised utterance and a 0.84 F-score drop is witnessed.
Since surface realisation only substitutes slot type with different slot values without
changing the utterances syntactically, this ablation shows it's more beneficial to
generate syntactic alternatives using our seq2seq model.

For our method without \textit{diversity ranks}, we remove diversity ranks from the utterance
representation and this lead a drop of 0.40 F-score.
We address the drop of performance to the fact that removing either these components
will lead to less diverse generation.
The second and third column in Table \ref{tbl:ablation} confirm this by 
showing less newly and diversely generated delexicalised utterances.

If we don't filter the alike instances when training the seq2seq model, the drop
of performance is a 0.65 F-score. However, larger number of new utterances
with smaller edit distances are yielded
which indicates that more noise is introduced when the training data of the seq2seq model is not properly
filtered.

This ablation also shows correlation between the maximum edit distance and the final F-score,
which indicates generating diverse augmentation helps the performance.

\paragraph{Effect of Training Data Size.}

\begin{table}[t]
	\centering
	\begin{tabular}{lrp{0.8\columnwidth}}
		\hline
		\multicolumn{3}{l}{\textit{show me all flights from atlanta to washington with prices}} \\
		\multicolumn{3}{l}{\textbf{(delex.)} \textit{show me all flights from} \texttt{<from\_city>} \textit{to} \texttt{<to\_city>} \textit{with prices}} \\
		\hline
		\#1 & train & let 's look at \texttt{<from\_city>} to \texttt{<to\_city>} again \\
		& ours & what are all the flights between \texttt{<from\_city>} and \texttt{<to\_city>} \\
		& & \textbf{(realized)} what are all the flights between indianapolis and tampa  \\
		\hdashline
		\#100 & train & list types of aircraft that fly between \texttt{<from\_city>} and \texttt{<to\_city>} \\
		& ours &  i 'm looking for a flight from \texttt{<from\_city>} to \texttt{<to\_city>} \\ 
		& & \textbf{(realized)} i 'm looking for a flight from milwaukee to los angeles \\
		\hdashline
		\multicolumn{2}{r}{Kurata16} &  show me all flights from [atlanta]\texttt{<from\_city>} to [washington]\texttt{<to\_city>} with airports\\
		\hline
		\hline
		\multicolumn{3}{l}{\textit{is there a flight between san francisco and boston with a stopover at dallas fort worth}} \\
		\multicolumn{3}{l}{\textbf{(delex.)} \textit{is there a flight between} \texttt{<from\_city>} \textit{and} \texttt{<to\_city>} \textit{with a stopover at} \texttt{<stop\_city>}} \\
		\hline
		\#1 & train & which airlines fly from \texttt{<from\_city>} to \texttt{<to\_city>} and have a stopover in \texttt{<stop\_city>} \\
		& ours & is there a flight from \texttt{<from\_city>} to\texttt{<to\_city>} with a stop in \texttt{<stop\_city>}  \\
		& &  \textbf{(realized)} is there a flight from washington to miami with a stop in dallas fort worth  \\
		\hdashline
		\#30 & train & do you have any airlines that would stop at \texttt{<stop\_city>} on the way from \texttt{<from\_city>} to \texttt{<to\_city>}\\
		& ours & i 'd like to fly from \texttt{<from\_city>} to \texttt{<to\_city>} with a stop in \texttt{<stop\_city>} \\ 
		& &  \textbf{(realized)} i 'd like to fly from memphis to boston with a stop in minneapolis  \\
		\hdashline
		\multicolumn{2}{r}{Kurata16} &  is there a flight between [san francisco]\texttt{<from\_city>} and [boston]\texttt{<to\_city>} with a stopover at [dallas fort worth]\texttt{<to\_city>}\\
		\hline
		
	\end{tabular}
	\caption{Case study of our augmented data against the training data and the results of \newcite{DBLP:conf/interspeech/KurataXZ16} (marked as Kurata16).
		\textit{train} marks the target utterance in the training data.
		\textbf{(delex.)} marks the delexicalised form of the input utterance.
		\textbf{(realized)} marks the utterance after surface realisation.
	}
	\label{tbl:case-study}
\end{table}

The results on ATIS and Stanford dialogue dataset witness the trend that
smaller training data benefits more from our augmentation method.
A natural question that arises is what's boundary of our augmentation
in the sense of improving the baseline.
In this section, we study this by varying training data size on the ATIS data.
Figure \ref{fig:data-vol} shows the results.
For the ATIS data, improvements can be achieved
in all our settings with training size smaller than one thousand.
These results indicate that our augmentation is applicable
when we only access to a LU training data of hundreds instances.

\paragraph{Case Study.}
In this paragraph, we perform case study on our method to
verify its capability of generating diversely augmented data.
Table \ref{tbl:case-study} shows two cases of our augmentation.
Each case
includes the original sentence and its delexicalised form (in \textit{italic} font),
the diversity rank (starts with \# mark), the training utterance under this rank,
our augmentation along with surface realization, and the augmentation
produced by \newcite{DBLP:conf/interspeech/KurataXZ16}.

By comparing our augmentation with the delexicalised form of source utterance,
two observations can be drawn: 1) our method yields syntactically different
alternatives meanwhile keeps the original semantic frame as the source utterance;
2) the lengths of the generated utterances are in the same scale with the source utterance
thanks to the effect of length penalty in Equation \ref{eq:score}.

By comparing our augmentation with the target training utterance under the same rank,
our seq2seq model yields different utterance instead of repeating the training utterance.
We address this diversity to the fact that our diversity rank has some
universal effect on modeling the diversity degree across different instances.
When contrasting to the augmentation of \newcite{DBLP:conf/interspeech/KurataXZ16},
our method clearly shows diverse augmentation against the source utterance while theirs are
basically repeating the source utterances. In the sense of generating diverse
alternatives for expressing the same semantics, our method has the advantage.

\section{Related work}

Data augmentation is an effective way of improving the model's performance
and it has been extensively explored on the computer vision community.
Single transformation approaches like randomly copying, flipping, and changing the intensity of RGB are
the common practice in the top-performed vision systems \cite{NIPS2012_4824}.
Beyond these classic approaches, adding noise to the image, randomly interpolating
a pair of images \cite{zhang2018mixup} are also proposed in previous works.
However, these signal transformation approaches are not directly applicable to language
because order of words in language may form rigorous syntactic and semantic meaning \cite{zhang2015character}.
Therefore, the best way of data augmentation in language usually involves generating the alternative expressions.

Paraphrasing is the most studied techniques in natural language processing for generating alternative
expressions \cite{barzilay-mckeown:2001:ACL,bannard-callisonburch:2005:ACL,callisonburch:2008:EMNLP}.
However, generic paraphrasing technique has been reported
not helpful for specific problem \cite{narayan-reddy-cohen:2016:INLG}.
Most of the successful work that applying paraphrasing for data augmentation
requires special tailored paraphrasing techniques.
For example, \newcite{wang2015s} performed word-level paraphrasing to extend their
corpus on twitter that contains annoying behaviors.
\newcite{fader-zettlemoyer-etzioni:2013:ACL2013} derived question templates
from seed paraphrases and bootstrap the templates to achieve the enlarged open-domain QA dataset.
\newcite{narayan-reddy-cohen:2016:INLG} constructed latent variable PCFG for questions
and augment the training data by sampling from the grammar.
All these works assume the same output (i.e. class in text classification, answer in question answering) for input paraphrases.
Our method resembles theirs in the assumption for input paraphrases, but differs
on using the seq2seq generation which is purely data-driven and doesn't rely on special tailored domain knowledge.
Besides these methods, works that introduce errors to language understanding have also been proposed \cite{schatzmann2007error,sagae2012hallucinated}.

Language understanding, as an important component in the task-oriented dialogue system pipeline,
has drawn a lot of research attention in recent year, especially when enhanced by the rich
representation power of the neural network, like recurrent neural network, LSTM \cite{yao2013recurrent,7078572,DBLP:conf/interspeech/MesnilHDB13,DBLP:journals/taslp/MesnilDYBDHHHTY15} and memory network \cite{contextualslu}.
Rich linguistic features \cite{chen2016syntax} and representation in broader scope on sentence-level \cite{kurata2016leveraging}
and dialogue history-level \cite{contextualslu} have also been studied.
Our augmentation method is orthogonal to these works and it's hopeful to achieve
more improvements with their works. 

Dialogue management is also a key component of task-oriented dialogue system, 
which mainly focuses on dialogue policy. 
However, optimal dialogue policy is hard to obtain from a static corpus
due to the vast space of conversation process. 
A solution is to transform the static corpus into user simulator \cite{kreyssig2018neural}, 
and most user simulators work on user semantics level.
\cite{658991,schatzmann2007agenda,asri2016sequence,scheffler2000probabilistic,scheffler2001corpus,pietquin2006probabilistic,georgila2005learning,cuayahuitl2005human}.
Recent work starts to generate user utterance directly to reduce data annotation\cite{kreyssig2018neural}.

In recent years, Generative Adversarial Network (GAN, Goodfellow et al. 2014)\nocite{NIPS2014_5423} 
draws a lot of research attention.
Its ability of generating adversarial examples is attractive for data augmentation.
However, it hasn't been tried in data augmentation beyond computer vision \cite{antoniou2018data}.
How to apply GAN to language understanding is still an open question. 

\section{Conclusion}

In this paper, we study the problem of data augmentation for LU.
We propose a data-driven framework to augment training data.
In our framework, one utterance's alternative expressions of the same semantic
are leveraged to train seq2seq model.
We also propose a novel diversity rank
to encourage diverse generation and filter alike instances.
In the experiments, our model achieves significant improvements
of 6.38 and 10.04 F-scores respectively when only a training set of hundreds utterances is represented.
Careful case study also shows the capability of our framework to generate diverse alternative expressions.
	
	

\section*{Acknowledgements}

We thank Xiaoming Shi for the LU annotation over the Stanford dialogue dataset.
We are grateful for helpful comments and suggestions from the anonymous reviewers.  
This work was supported by the National Key Basic Research Program of China via grant 2014CB340503 
and the National Natural Science Foundation of China (NSFC) via grant 61632011 and 61772153.

\bibliographystyle{acl}
\bibliography{coling2018}

\end{document}